\documentclass[letterpaper, 10 pt, conference]{ieeeconf}  %

\IEEEoverridecommandlockouts                              

\usepackage{graphicx} 
\usepackage{amsmath} 
\usepackage{amssymb}  
\usepackage[usenames, dvipsnames]{color}

\usepackage{lipsum}
\usepackage{color}
\usepackage{cite}

\usepackage{diagbox} 
\usepackage{tabu}
\usepackage{balance}  

\usepackage[ruled,linesnumbered]{algorithm2e}
\usepackage{algpseudocode}
\usepackage{epsfig}

\usepackage{booktabs,tabulary,lipsum}
\usepackage{dcolumn,tipa}
\newcolumntype{d}{D{.}{.}{6.5}}
\usepackage{siunitx}
\usepackage[usestackEOL]{stackengine}
\usepackage{multirow}
\usepackage{makecell}

\newcommand{\revision}[1]{\textcolor{black}{#1}} 

\title{

\revision{Tactile-Guided Robotic Ultrasound:\\Mapping Preplanned Scan Paths for Intercostal Imaging}
}

\author{Yifan Zhang, Dianye Huang, Nassir Navab, \textit{Fellow, IEEE}, and Zhongliang Jiang 
\thanks{This work was supported in part by the Multi-Scale Medical Robotics Center, AIR@InnoHK, Hong Kong; and in part by the SINO-German Mobility Project under Grant M0221.}
\thanks{$^{1}$ Yifan Zhang, Dianye Huang, Nassir Navab and Zhongliang Jiang are with the Chair for Computer Aided Medical Procedures and Augmented Reality, Technical University of Munich, Germany. D. Huang, N. Navab, and Z. Jiang are also affiliated with the Munich Center for Machine Learning (MCML), Munich, Germany. {\tt\footnotesize{(zl.jiang@tum.de)}}
}%
}

\begin{document}

\maketitle


\begin{abstract}
Medical ultrasound (US) imaging is widely used in clinical examinations due to its portability, real-time capability, and radiation-free nature. To address inter- and intra-operator variability, robotic ultrasound systems have gained increasing attention. However, their application in challenging intercostal imaging remains limited due to the lack of an effective scan path generation method within the constrained acoustic window. To overcome this challenge, we explore the potential of tactile cues for characterizing subcutaneous rib structures as an alternative signal for ultrasound segmentation-free bone surface point cloud extraction. Compared to 2D US images, 1D tactile-related signals offer higher processing efficiency and are less susceptible to acoustic noise and artifacts. By leveraging robotic tracking data, a sparse tactile point cloud is generated through a few scans along the rib, mimicking human palpation. To robustly map the scanning trajectory into the intercostal space, the sparse tactile bone location point cloud is first interpolated to form a denser representation. This refined point cloud is then registered to an image-based dense bone surface point cloud, enabling accurate scan path mapping for individual patients. Additionally, to ensure full coverage of the object of interest, we introduce an automated tilt angle adjustment method to visualize structures beneath the bone. To validate the proposed method, we conducted comprehensive experiments on four distinct phantoms. The final scanning waypoint mapping achieved \revision{Mean Nearest Neighbor Distance (MNND)} and \revision{Hausdorff distance (HD)} errors of 3.41 mm and 3.65 mm, respectively, while the reconstructed object beneath the bone had errors of 0.69 mm and 2.2 mm compared to the CT ground truth. 
\end{abstract}



\bstctlcite{IEEEexample:BSTcontrol}
\section{Introduction} \label{sec:intro}
\par
Medical ultrasound (US) has been widely used for examining internal organs, such as the liver and heart. Compared to imaging modalities like CT and MRI, US is versatile, accessible, and cost-effective. Additionally, due to the advantage of real-time imaging, the US is also commonly used to develop image-guided intervention procedures~\cite{huang2025vibnet} in practice, such as liver ablation~\cite{tsang2021high}. However, US image quality is highly sensitive to machine variations, imaging parameter settings, and acquisition factors such as probe pressure and orientation. To address inter- and intra-operator variability, robotic systems have been introduced to improve scanning precision and reproducibility~\cite{jiang2023robotic, li2021overview,von2021medical, bi2024machine, chen2025uspilot}. 

\begin{figure}[t!]
\centering
\includegraphics[width=0.95\columnwidth]{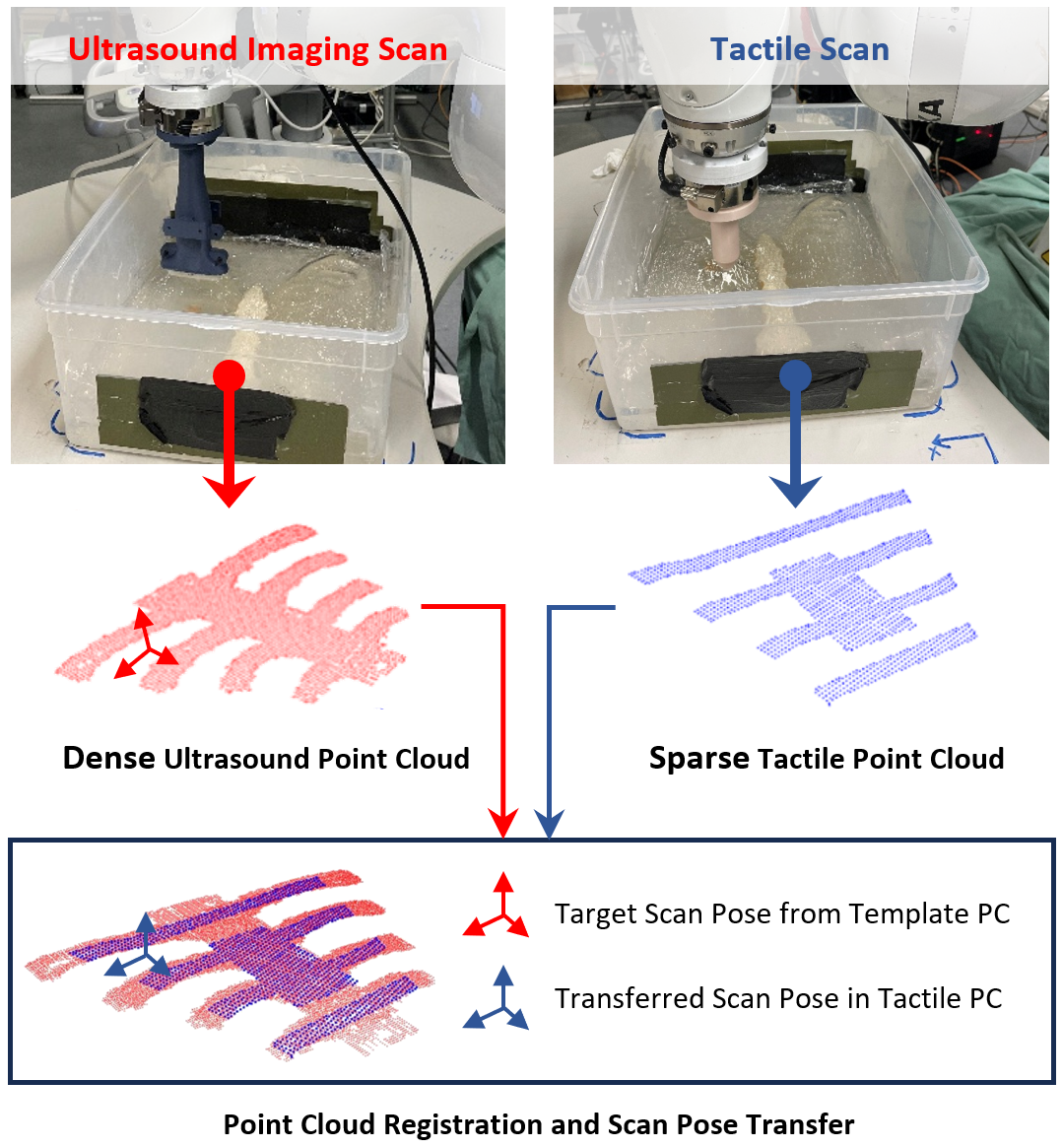}  
\caption{An illustration of transferring the preplanned scan pose from the dense US point cloud template (\textit{Left}) onto the onsite acquired sparse tactile point cloud (\textit{Right}) extracted from the tactile data.
}
\label{fig:teaser}
\end{figure}

\par
Recently, various autonomous and semi-autonomous robotic ultrasound systems (RUSS) have been developed for diverse applications, including vascular~\cite{jiang2024intelligent, ning2023autonomous, bi2025gaze, huangqinghua2024robot, li2025semantic, huang2024robot, dall2024imitation}, heart~\cite{zakeri2025robust}, spinal~\cite{ma2025cross, duan2022ultrasound}, lung~\cite{ma2024guiding}, breast~\cite{tan2022flexible, huang2023mimicking}, and liver imaging~\cite{lei2023robotic}. To ensure successful execution of these scanning tasks, a task-specific scanning path must be computed and tailored to the requirements of each application. In order to plan such a path for vascular structure, Jiang~\emph{et al.} presented an online path optimization solution using real-time segmentation masks of vascular US images~\cite{jiang2021autonomous}. However, preplanning on pre-operative CT/MRI or camera-based object surfaces is dominating. Huang~\emph{et al.} computed the multiple-line scan path based on the captured RGB image of the abdominal surface~\cite{huang2018robotic}. To have a precise 3D representation, Tan~\emph{et al.} performed scanning path planning using fused body surface point clouds from multiple viewpoints~\cite{tan2022fully}. 

\par
In order to map the scanning trajectory to individual patients for US examination, the registration between the current scene and preoperative imaging coordination frame should be computed. Hennersperger~\emph{et al.} present a method to match the skin surface point clouds from a live camera and a template MRI~\cite{hennersperger2016towards}. Considering inter-patient variations, Jiang~\emph{et al.} used non-rigid registration to generate patient-specific scanning paths by explicitly considering the articulated motion around limb joint~\cite{jiang2022towards}. Although these approaches have shown promising results in their respective applications, their direct extension to tasks requiring high precision may be challenging. For instance, mapping a scanning path through the narrow intercostal space for liver or cardiac imaging presents significant difficulties~\cite{bi2024autonomous}. In such cases, the ribs, with their high acoustic impedance, create acoustic shadows that obscure the visibility of underlying tissues, potentially compromising imaging quality. To tackle this challenge, Jiang~\emph{et al.} proposed a skeleton-graph-based registration method, which directly uses the subcutaneous rib cage surface to ensure the path can be in between the rib bones~\cite{jiang2024class, jiang2023skeleton, jiang2023thoracic}. Their results demonstrate the effectiveness of ensuring an optimal intercostal imaging view. However, the reliance on both bone surface segmentation and classification limits its practicality and effectiveness in clinical applications.

\par
Drawing inspiration from palpation, the force variation experienced when running a finger across the chest suggests the potential for leveraging tactile cues. Therefore, we explore the integration of tactile feedback to enhance the detection and characterization of individual rib cage features. In the previous study, the tactile cue in the scanning of RUSS has been used for tumor localization~\cite{naidu2017breakthrough, beber2024passive}, US image deformation correction~\cite{jiang2021deformation, jiang2023defcor}, and computing static US elastography~\cite{patlan2016automatic}. 

\par
In this study, we first explore the potential of using tactile cues as an alternative signal for ultrasound-free bone surface point cloud extraction. After interpolation, a denser tactile surface point cloud is created and is further utilized in the registration algorithm to align the preoperative scanning path with the current setting. Compared to 2D images, processing 1D tactile signals is more efficient and less susceptible to acoustic noise and artifacts inherent in US images. To the best of our knowledge, this is the first use of tactile information over 2D US images to characterize subcutaneous bone structure for precise scan path mapping. To validate our method, we evaluated bone detection performance on three distinct phantoms using the same model and assessed the challenging intercostal path mapping performance on a custom-made rib cage phantom with a piece of \revision{porcine tissue} beneath the bone. \revision{To avoid potential misunderstanding, we would like to emphasize that we do not claim that relying only on the tactile cue is superior to using the US image itself for characterizing bone structure.} An intuitive demonstration can be found in this online video\footnote{Video: https://youtu.be/SBwpFVzEhAg}.

\par
The rest of this paper is organized as follows. Section \ref{sec:pre} introduces hardware setup and the preparation of the template point cloud and tactile data. Section \ref{sec:method} describes the overall structure of the proposed method. The experimental results are provided and discussed in Section \ref{sec:exp}. Finally, the summary of this study is presented in Section \ref{sec:conclusion}. 


\begin{figure*}[t!]
\centering
\includegraphics[width=0.8\textwidth]{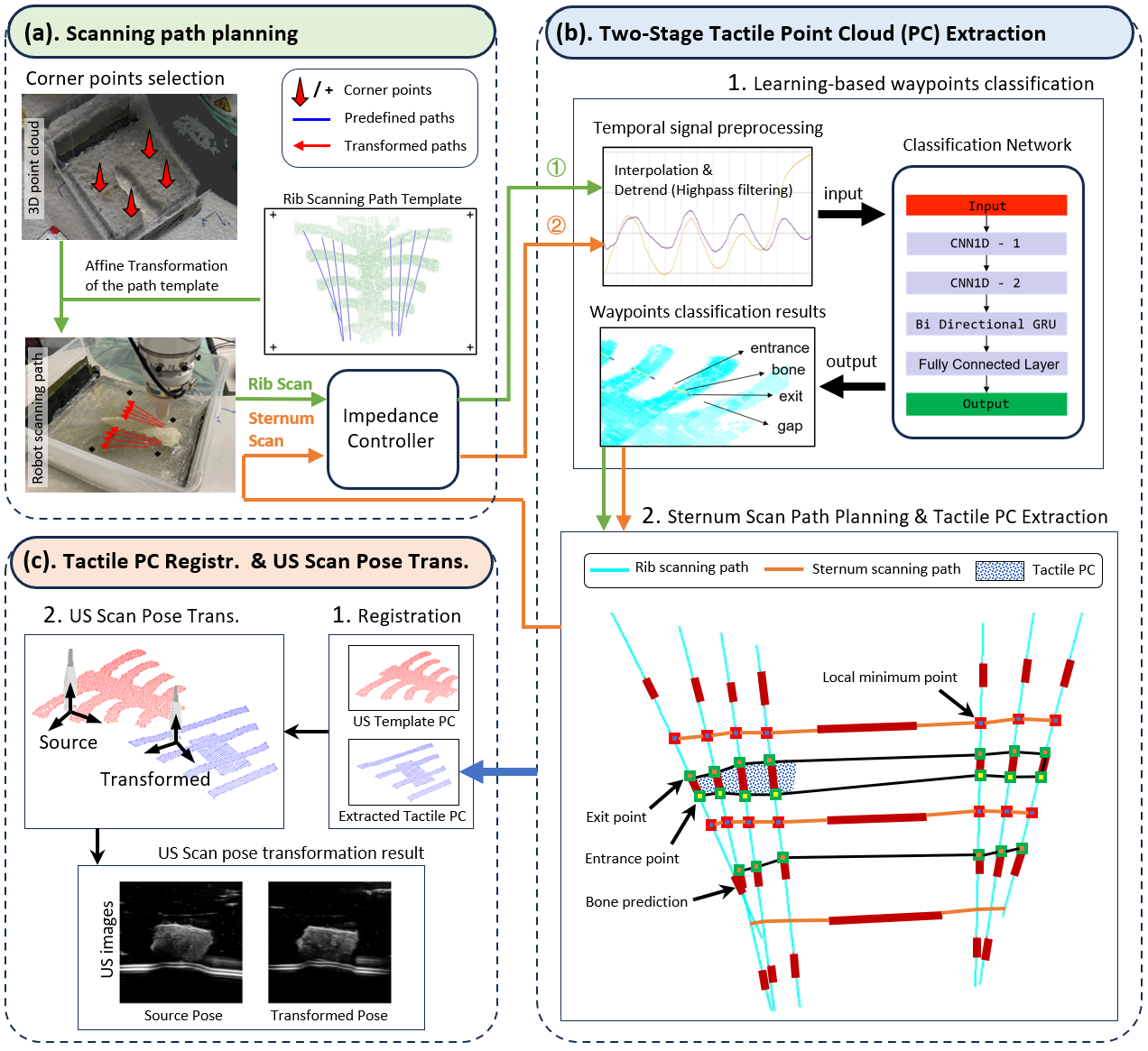}  
\caption{A system overview illustrating the processes of tactile point cloud registration and US scan pose transformation. 
}
\label{fig:overview}
\end{figure*}

\section{Preliminaries}
\label{sec:pre}
\subsection{Hardware Setup} \label{sec:pre-I}
\par
To mimic the intercostal scanning scenario, we use a rib bone from a commercial human skeleton model (Hadwyn) embedded in candle wax gelatin, constructed layer by layer. To assess the effectiveness and robustness of the tactile-based bone detection algorithm (see \textit{Section} \ref{subsec:bone_detec}), the final layer covering the ribs varies in thickness: 7 mm for chest phantom A and 10 mm for chest phantom B. Due to the use of candle wax gelatin, \revision{hantoms A and B can have a soft skin layer with a stiff bone structure beneath.} Since human tissue exhibits different stiffness properties, we further evaluate the bone detection algorithm on the ribs of a stiff commercial abdominal phantom (US-22, Kyoto Kagaku Co., Ltd., Japan) to demonstrate the method can be applied to objects with varying stiffness. To fully validate the pipeline, chest phantom C is constructed similarly but incorporates \revision{porcine tissue} to better simulate the region of interest beneath the rib cage. 
Notably, the embedded rib bone for phantoms A, B and C are identical.

\par
As shown in Fig. \ref{fig:teaser}, the experimental setup consists of a collaborative robotic arm (LBR iiwa 7 R800, KUKA GmbH, Germany) equipped with a 6-axis force/torque sensor (GAMMA, ATI, USA) for precise force measurement. To obtain a complete surface point cloud representation of the rib bone, a one-time US scan was performed on phantom A, followed by manual annotation of the rib bone surface using ImFusion software. The US images were captured by a linear probe (12L3, Siemens, Germany) connected to a US machine (ACUSON Juniper, SIEMENS AG) via a frame grabber (USB Capture HDMI, MAGEWELL, China). To obtain clear bone structures in the B-mode images, we applied the default settings provided by the manufacturer: MI: $1.13$, TIS: $0.2$  TIB: $0.2$ DB: 60 dB. Given the shallow depth of the target ribs, the imaging depth was set to $40~mm$. Leveraging robotic tracking data from the manipulator, we can reconstruct a full 3D point cloud of the subcutaneous bone surface.

\par
To acquire the tactile point cloud (PC) characterizing the rib bone, a 3D-printed \textit{Indenter}—a plastic cone with a spherical tip, as shown in Fig. \ref{fig:teaser}—is mounted on the robotic flange. Using the 3D model of the \textit{Indenter}, the tip of the tactile tool is tracked via the robot’s forward kinematics. Notably, tactile scanning is performed sparsely to optimize time efficiency. Given the structural similarity of human chests, we define eight scan lines parallel to the sternum centerline and three scan lines orthogonal to it for structured data acquisition. To approximate the mapping of sparse scanning lines on gelatin phantoms, an RGB-D sensor (Azure Kinect, Microsoft, USA) is used. By selecting four corner points, we define the region of interest for alignment. By stacking the bone-identified locations tracked by the robot, we obtain a sparse tactile PC. This sparse PC is then interpolated to generate a dense tactile PC, which is used to map the scan path based on the registration results.

\subsection{Data Preparation}

\subsubsection{Temporal Calibration between US Stream and Tracking Stream}
To extract a reliable PC from the US and tactile data, it is crucial to perform temporal calibration to synchronize the system's sensor signals, including the tracking data from the robotic arm, the US image data, and the force sensor data. \revision{It is worth noting that the US PC (manually annotated) is only used here to provide a preplanned path and for quantitative evaluation for path mapping.} Similar to the method outlined in~\cite{salehi2017precise}, we synchronize the force sensor data with the tracking data by commanding the robotic arm, equipped with an \textit{Indenter}, to periodically interact with a chest phantom (which will be introduced later). We then extract the peaks and troughs of both periodic signals and calculate the temporal difference by aligning their corresponding peaks and troughs. Similarly, we command the robotic arm with a US probe to periodically and vertically move within a water tank containing an object. The temporal difference is computed by synchronizing the periodic changes in the position of the object in the US images with tracking data. Finally, in our setup, the temporal differences between the US image data and the force sensor data, relative to the robotic arm's tracking data, are $268~ms$ and $32~ms$, respectively. In addition, the spatial calibration between the US image and the robotic base frame can refer to~\cite{jiang2021autonomous}. 

\subsubsection{Template Point Cloud from US Sweep}
\begin{figure}[h]
\centering
\includegraphics[width=\columnwidth]{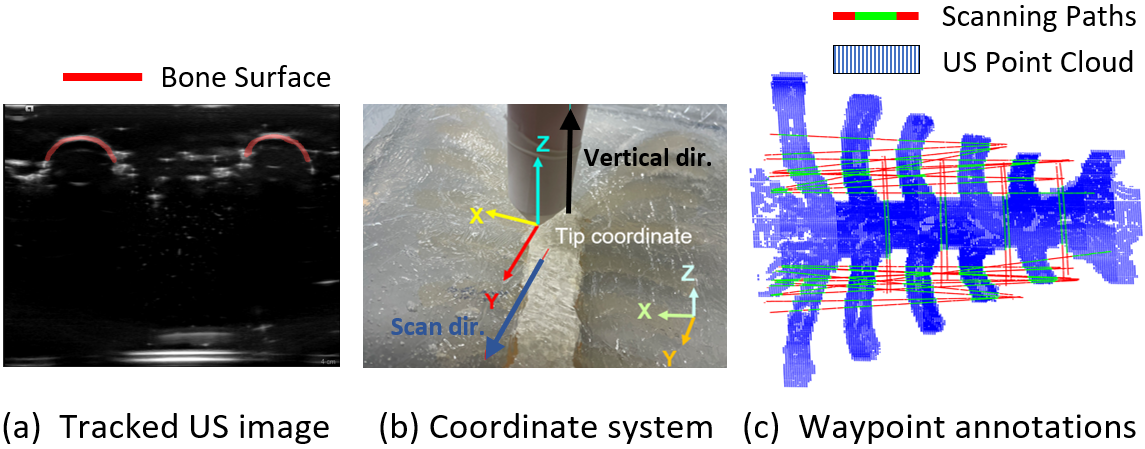}  
\caption{An illustration of (a) bone surface annotation in a tracked US image, (b). coordinate system of the tactile scan, and (c). the reconstructed bone surface PC, with annotated waypoints along the tactile scanning paths. Green line segments represent bone surfaces, while red line segments indicate non-bone surfaces.}
\label{fig:annotation}
\end{figure}

A template PC is needed to preplan the scanning path and map this path onto the onsite acquired tactile PC. \revision{In this study, we create the US PC by manually annotating the bone surface from tracked US images.} To do so, we build chest phantoms (see Sec.~\ref{sec:pre-I}) using gelatin. Inside phantom C, a piece of \revision{porcine tissue}, simulating the scanning target, is embedded beneath a 3D-printed, human-like rib structure. The robotic arm equipped with a linear US probe at its end-effector scans the human-like rib phantom with a constant contact force of 3.0 $N$, which is in the safe range for the majority of the US examination~\cite{jiang2023robotic}. 
Multiple scans are performed to ensure full coverage of rib structure. During the scan, the US images and the robotic arm's end-effector poses are recorded simultaneously and synchronized. Subsequently, the bone surfaces of the tracked US images, including the ribs and sternum, are manually labeled [see Fig. \ref{fig:annotation} (a)]. The 3D reconstruction of the bone surface [see Fig. \ref{fig:annotation} (b)] is then performed using the labeled tracked US images\footnote{The 3D reconstruction is done with the \textit{ImFusion Suite} (ImFusion GmH, Munich, Germany)}. It is noted that since the human chest is almost flat, we remove the upward direction of the 3D PC so the registration can be performed in 2D space. Finally, we down-sample the 2D PC to obtain a template PC with uniformly distributed point elements. This bone surface PC will serve as the US template PC in the subsequent process.

\subsubsection{Tactile Data Acquisition and Annotation}
\label{subsec:annotation}
The tactile data are collected to train a neural network for tactile PC extraction. To achieve this, the robotic arm, equipped with an \textit{Indenter}, is commanded to perform sweep scans on the chest phantom. As depicted in Fig. \ref{fig:annotation} (b), the \textit{Indenter} follows a predefined scan path at a constant speed of $4.86~mm/s$ while applying a constant contact force of 3.0 N. 
During the scan, 
the poses of the \textit{Indenter}’s tooltip is recorded.
A total of 60 scans are performed by randomly positioning the chest phantom at 4 different locations with varying heights, ensuring the diversity of the collected data. It is noted that all the scanning paths are straight lines.

\par
For each scan, we collected a series of waypoints.
A system with tactile sensing capability should be able to distinguish between the bone and the non-bone surfaces based on the tactile data. Notably, before acquiring the tactile data, a US imaging scan is conducted to generate a template US PC for annotating the tactile data. Consequently, each waypoint in the tactile data can be labeled as ``bone'' [see green line in Fig. \ref{fig:annotation}(c)] if it falls within the bone region of the US template PC; otherwise, it is labeled as ``gap'' [see red line in Fig. \ref{fig:annotation}(c)]. As the robot continuously performs the sweep scan, the system is also expected to identify the transition points where the \textit{Intender} makes contact with the bone from a non-bone area and vice versa. Hence, we further introduce two more classes, ``entrance" and ``exit", to the ground truth labels, which correspond to the transition point between ``bone" and ``gap" regions [see the ``Entrance'' and ``Exit'' points in Fig. \ref{fig:overview} (b-2)]. 
\section{Methods}
\label{sec:method}
In this section, we detail the three modules that constitute the proposed system: (1) the scanning path planning module for raw tactile data acquisition, (2) two-stage tactile point cloud extraction, and (3) tactile point cloud registration that enables the preplanned US pose transfer. An overview of the proposed system is provided in Fig. \ref{fig:overview}.

\subsection{Scan Path Planning for Tactile PC Generation}
To roughly define the chest area of the phantom, we use an RGB-D camera to perceive the workspace and manually select four corner points in RViz, a 3D visualization tool in Robot Operating System (ROS) that allows users to interactively inspect and manipulate sensor data, robotic states, and environment representations. These 3d corner points, along with a predefined 2D rib scanning path template [see Fig. \ref{fig:overview} (a)], can enable us to roughly map multiple scan paths onto the chest phantom's surface in the workspace.

\par
To map the ribs scanning path template, a 2D texture containing multiple predefined scanning paths, onto a 3D chest phantom's surface, we first need to define a projection plane in 3D. This can be formulated as finding a plane that minimizes its accumulated distance to the selected corner points:
\begin{equation}
    \begin{aligned}
        \pi^* = \arg \min_{a,b,c,d} \quad & \sum_{i=1}^{4} \left|\left| ax_i + by_i + cz_i + d \right|\right|^2 \\
        \text{s.t.} \quad & a^2 + b^2 + c^2 = 1.
    \end{aligned}
\label{eq:plane_optim}
\end{equation}
where $\pi^*=\{(x, y, z)\in\mathbb{R}^3|ax+by+cz+d=0\}$ is the optimal plane that best fits the four selected corner points. We employ the Principal Component \revision{Analysis} (PCA) to solve the optimization problem formulated in Eq. (\ref{eq:plane_optim}). Subsequently, we compute new corner points $\mathbf{p}^\prime$ by projecting each selected corner point $\mathbf{p}$ onto $\pi^*$ using $\mathbf{p}^\prime=\mathbf{p}-d_{\perp}\mathbf{n}$, where $d_{\perp}$ denotes the distance between $\mathbf{p}$ and $\pi^*$, and $\mathbf{n}$ is the normal vector of $\pi^*$. As a result, we obtain a projection plane $\pi^*$ with four adjusted corner points $\mathbf{p}^\prime$ in 3D space. With these updated 3D points and the 2D corner pixel coordinates from the scanning path templates, we can further obtain the 2D-to-3D affine transformation by optimizing the following equation:
\begin{equation}
    \mathbf{T}^* = \arg \min_{\mathbf{T}} \quad \sum_{i=1}^{4} \left|\left| \mathbf{p}_{i}^{\prime} - \mathbf{T}\cdot\mathbf{p}_{i}^{2D}\right|\right|^2
    \label{eq:affine_optim}
\end{equation}
where $\mathbf{T}^*\in\mathbb{R}^{3\times3}$ is the optimal affine transformation matrix $\mathbf{p}_{i}^{2D}=[u,~v,~1]^T$ represents the homogeneous 2D coordinates of the scanning path template. Using $\mathbf{T}^*$, we can map the predefined 2D scanning path onto the 3D surface of the chest phantom [see Fig. \ref{fig:overview} (a)] and the z-axis of the \textit{Intender} [see Fig. \ref{fig:annotation} (b)] aligns with the normal vector $\mathbf{n}$.

Notably, to obtain a complete tactile PC that includes the sternum, additional scan paths are required. To achieve this, the rib's tactile PC is first extracted by classifying waypoints along the scan path using a lightweight network, as introduced in \textit{Section} \ref{subsec:network}. As illustrated in Fig. \ref{fig:overview} (b-2), local minima between the ribs are detected and connected to generate three additional scan paths for identifying the sternum. The detailed extraction process is presented in the following subsection.

\par
During the scanning, the robotic is controlled in impedance mode ~\cite{hennersperger2016towards} to ensure safe contact with objects. The Cartesian compliant control law is defined as below: 

\begin{equation}\label{eq_impedance_law}
\tau = J^{T}[K_m(x_d - x_c) + F_d] + D(d_m) + f_{dyn}(q,\Dot{q}, \Ddot{q})
\end{equation}
where $\tau$ is the target torque applied in the joint space, $J^{T}$ is the transposed Jacobian matrix, $x_d$ and $F_d$ are the desired position and desired force, $x_d$ is current position, $K_m$ is the given Cartesian stiffness, $D(d_m)$ represents the damping term, and $f_{dyn}(q,\Dot{q}, \Ddot{q})$ represents dynamics. 

\par
The impedance controller behaves like a spring with stiffness \( K_m \). If an unexpected obstacle, such as a patient's body part, obstructs the path, the robot halts at the contact point. However, if the resistance force is insufficient, the manipulator continues toward the target. To ensure compliance, the stiffness along the probe centerline is set to 1 N/m, while the other two directions are maintained at 500 N/m to keep the probe aligned with the scanning path.

\subsection{Scanning Waypoints Classification}
\label{subsec:bone_detec}
After mapping the template paths to the workspace, the robotic arm is commanded to perform the tactile scan, tracking multiple scan paths while collecting the poses of the attached \textit{Indenter}.

\subsubsection{Signal Preprocessing}
In this work, we employ impedance control to regulate the interaction between the \textit{Indenter} and the chest phantom. To maintain a constant contact force along the scan path, the equilibrium point of the impedance controller is continuously adjusted as the \textit{Indenter} transitions between the ``gap" and ``bone" regions. This allows us to leverage variations in indentation depth—i.e., the z-axis displacement $\Delta z$ of the \textit{Indenter} [see Fig. \ref{fig:annotation}(b)]—to extract implicit tactile information. To enhance compliance and amplify variations in the equilibrium point, we deliberately set the stiffness of the z-axis (indentation direction) to 1 N/m. To ensure a uniform distribution of points in the tactile point cloud, we resample the tracking data of the \textit{Indenter}. Additionally, to minimize interference from the uneven surface of the chest phantom, we apply a high-pass filter to the $z$-position before computing the displacement $\Delta z$. This filtering removes wavelength components larger than the typical bone-to-bone distance (approximately $2.5~cm$), preserving only the fluctuations caused by the presence of bones and gaps.

\subsubsection{Network Structure}
\label{subsec:network}
A lightweight recurrent network is designed to achieve the classification task. This network accepts a time-series signal $\Delta z_{400}$ of length 400 as input. The input is processed through a series of 1D CNN layers, GRU layers, and fully connected layers [see Fig. \ref{fig:overview} (b)]. The 1D CNN layers are responsible for extracting local signal features, while the GRU layers capture longer-range temporal dependencies between frames. The network produces an output with 4 channels over 400 frames, where each channel corresponds to the probability that a given element belongs to one of the classes (``bone", ``gap", ``entrance", or ``exit" defined in \textit{Section} \ref{subsec:annotation}). A softmax activation is applied to the output, and the network is trained using cross-entropy loss. During prediction and segmentation, we focus on the ``bone" channel: if the softmax value exceeds 0.9, the frame is classified as ``bone''; otherwise, it is classified as a ``gap''.



\subsection{Tactile Point Cloud Generation and Registration} \label{sec:method_registration}
In order to form closed areas from the classified points on the scanning path, it is necessary to determine the waypoints on different scanning paths that belong to the same rib. We firstly use PCA to find the main direction of each scanning path, the normal vector $\mathbf{v_i}$ and find a $\mathbf{v_{mean}} = \frac{1}{n} \sum_{i=1}^{n} \mathbf{v}_i$. After that, we project the waypoints on every scanning path that is classified as ``bone" to this direction ${value_j}=\mathbf{p_j}\cdot\mathbf{v_{mean}}$. Because the scanning lines are basically perpendicular to the rib's direction, this ${value_j}$ represents how far the point's relative distance to the rib is. Hence, we can classify the points belonging to the same ribs by classifying the points' projection ${value_j}$ that are clustering together using the density-based spatial clustering of applications with noise (DBSCAN) algorithm. 


\par
Notably, classification is only required for the boundary points along each scanning path. By connecting boundary points belonging to the same cluster, we can delineate the rib boundaries. The interpolation of the PC is carried out in two steps: i) Intra-rib interpolation: Boundary points belonging to the same rib are interpolated to create a smooth and continuous representation.
ii) Inter-boundary interpolation: Corresponding points across different boundary lines are interpolated to form a structured point cloud. Following this process, the rib area is effectively filled with interpolated points, generating a complete tactile point cloud. The sternum’s point cloud is constructed in a similar manner.



\par
After obtaining the tactile point cloud, we apply the Coherent Point Drift (CPD) algorithm \cite{myronenko2010point} to register it with the template ultrasound point cloud. CPD formulates point cloud registration as a probability density estimation problem and solves it using the Expectation-Maximization (EM) algorithm. This approach is particularly well-suited for our task due to its robustness against noise.

\subsection{Scan Path Generation for Intercostal Imaging}

\begin{figure*}[h]
\centering
\includegraphics[width=0.8\textwidth]{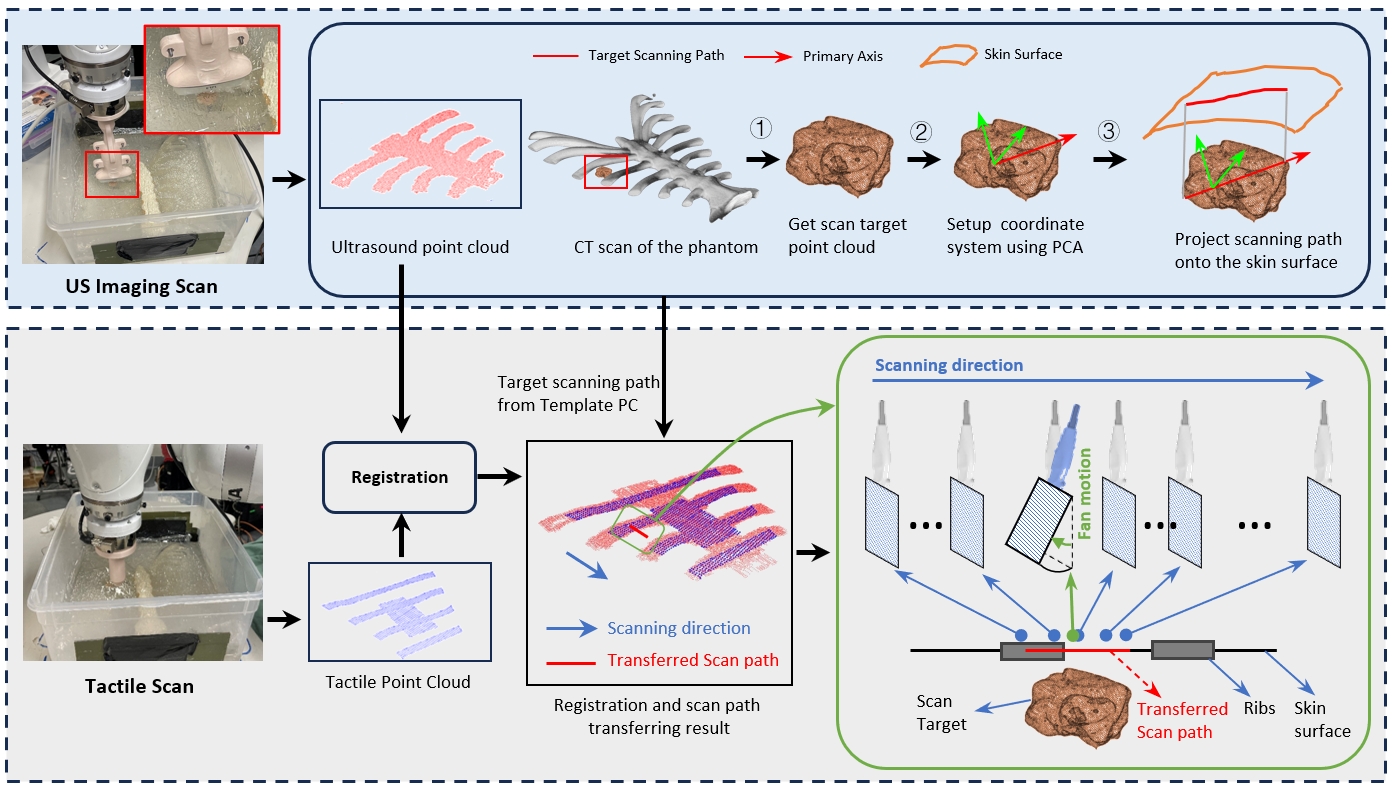}  
\caption{An illustration of \textit{Upper row:} generating the intercostal scanning path in the template PC, and \textit{Lower row:} robotic scan trajectory adjustment for ROI coverage when tracking the transferred scanning path.}
\label{fig:waypoint_reg}
\end{figure*}


\subsubsection{Scanning Path Generation}
\par
The intercostal scanning path is automatically generated from a pre-scan ultrasound sweep on Phantom C, with \revision{the porcine tissue} beneath the rib. Notably, pre-recorded US volume is adopted here for planning the examination scanning path. But it can be readily replaced by other imaging modalities, such as CT or MRI~\cite{jiang2024class}. 
In our setup, the meat in the gelatin phantom appears clearly in the ultrasound images [see Fig.~\ref{fig:overview} (c)] and can be easily segmented using Otsu’s thresholding. We then compute the centroid of the segmented target in each 2D slice. Utilizing spatial calibration and robotic tracking data, we obtain a point set \( \textbf{P}_{obj} \), representing the centroids of all frames containing meat in the template image space. To optimize the scanning path, PCA is applied to \( \textbf{P}_{obj} \), determining the optimal trajectory in the template space. The computed primary axis, closely aligned with the sternum centerline, is chosen as the scanning direction, with the start and stop points determined based on the centroid's projection along this axis (see the top section of \revision{Fig.~\ref{fig:waypoint_reg}}). While the current scanning path is derived from preoperative ultrasound scans, it can be adapted to other modalities, such as CT or MRI, depending on the application.


\subsubsection{Robotic Scan Trajectory Adjustment for Full ROI Coverage}
\par
Based on the registration between the full ultrasound-based PC and the tactile PC (see Fig.~\ref{fig:waypoint_reg}), the scanning trajectory is mapped from the template PC space to the current tactile PC setting. The default probe orientation is normal to the contact surface. However, due to registration errors or bone-induced acoustic shadows, further adjustments may be necessary to ensure full coverage of the region of interest ROI. During scanning, the US image is segmented in real time using Otsu’s thresholding. If the bone surface (typically located at the top of the image) transitions to the ROI, or vice versa, it indicates that part of the meat is obstructed by the rib. To address this, a fan motion is triggered to capture the missing object beneath the bone. To achieve this, the probe is first moved inward by a distance vector \( L_{adj} \), where the sign is positive if the movement aligns with the scanning direction and negative otherwise. The tilt angle $\theta_{adj}$ from the probe centerline is then determined based on the object's depth in the ultrasound image at the nearest intercostal space to the rib bone.

\begin{equation}\label{eq_adjustment}
\theta_{adj} = sgn(L_{adj})~\text{arctan}(\frac{L_{adj}}{c_{h} s_h})
\end{equation}
where \( c_h \) is the centroid height (in pixels) from the top of the segmented US image, and \( s_h = 0.067\) is the scaling factor for mapping pixels to millimeters. Since the object is located at approximately 20 mm depth in the image, \( L_{adj} \) is empirically set to 10 mm to minimize excessive probe tilting in this study.

\section{Experiments and Results}
\label{sec:exp}
\subsection{Tactile-based Bone Detection Performance}
\par
Due to the low stiffness along the probe centerline during impedance control with a constant desired force, variations in interaction force are reflected in positional changes, which can be precisely measured through robotic kinematics. To detect the bone location, the preprocessed 1D displacement signal is used as input for a simple bone localization network. The model was trained on 60 tactile scanning lines from gelatin phantom A. To ensure a fair comparison and assess the robustness of tactile cue-based bone localization, we tested the trained model on 10 unseen scanning lines from phantom A, along with an additional 20 lines from unseen phantom B and the abdominal US-22 phantom (10 from each). Accuracy is defined as the percentage of predicted ``bone" pixels that fall within the labeled ``bone" region. Distance shift represents the distance between the predicted ``bone" section's centroid and the ground truth centroid. The prediction accuracy was 91.15\% for phantom A and 88.16\% for phantom B. While a slight decrease in performance was observed for phantom B, the centroid shift of the detected bone was only 1.36 mm, which remains within an acceptable range. As shown in Tab.~\ref{segmentation_accuracy}, the detection accuracy for the commercial abdominal phantom with high stiffness decreases to 78\%. However, considering the average intercostal gap of approximately 30 mm, a 3 mm positional shift still allows the probe to be guided to the intercostal space for imaging with minimal or no acoustic shadowing.


\begin{table}[ht]
\centering
\sisetup{
    table-number-alignment = center,
    table-figures-integer = 1,
    table-figures-decimal = 4
}
\caption{Bone Detection Accuracy and Centroid Shift}
\label{segmentation_accuracy}
\renewcommand\footnoterule{\kern -1ex}
\renewcommand{\arraystretch}{1.3}
\resizebox{0.48\textwidth}{!}{
    \begin{tabular}{lcc}
        \toprule
        \textbf{Phantoms} & \textbf{Accuracy (\%)} & \textbf{Distance Shift (mm)} \\ 
        \midrule
        Gelatin Phantom A (7mm) & 91.15 ± 5.70 & 1.13 \\ 
        Gelatin Phantom B (10mm) & 88.16 ± 7.47 & 1.36 \\ 
        Abdominal US-22 phantom & 79.81 ± 14.14 & 3.31 \\ 
        \bottomrule
    \end{tabular}
}
\end{table}

\begin{figure}[h]
\centering
\includegraphics[width=0.7\columnwidth]{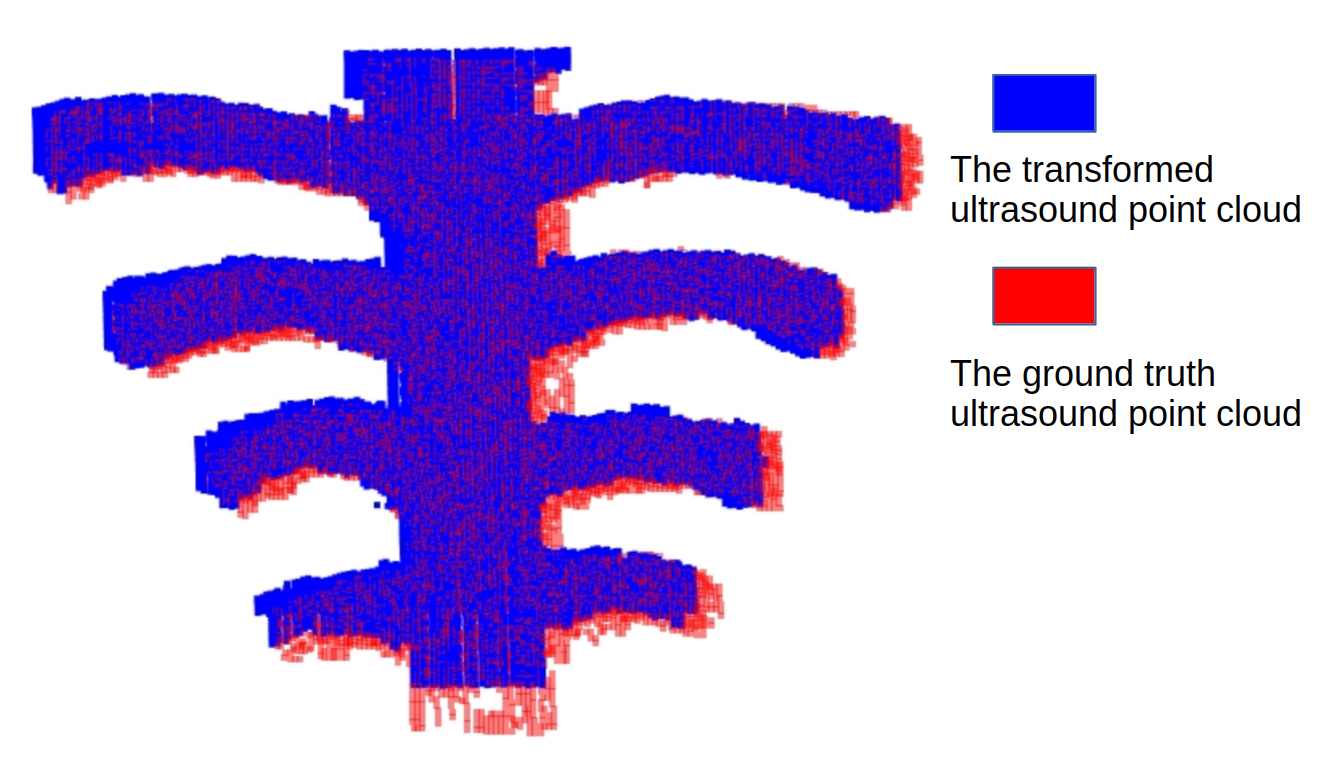}  
\caption{The ultrasound ground truth point cloud and inversely transformed ultrasound point cloud}
\label{fig:regis}
\end{figure}

\subsection{Registration Performance}
In this section, we present the registration results computed between the tactile PC and the full-level US PC created by manual annotation. To evaluate the registration accuracy, we manually altered the gelatin phantom’s position from its original location during the US scans. These positional changes were applied randomly in both rotational and translational directions. To quantify translational and rotational errors, we first computed the registration matrix between the full rib surface point cloud obtained at two random locations and the US PC template, which serves as the ground truth. We then compared the registration results computed between the reference US PC and the sparse tactile PCs at these two distinct locations directly against the ground truth.

The results in Tab.~\ref{table_registration_results} show that the translational errors for the two trials are 3.38 mm and 3.58 mm, while both rotational errors remain below one degree. These findings demonstrate that even a sparse tactile PC can achieve precise point cloud registration. To intuitively demonstrate the registration performance, we map the full-level US PC collected at location 1 to the template PC space. The overlaid two dense PCs is intuitively shown in Fig.~\ref{fig:regis}.

\par
To further assess the method's sensitivity to tactile scan path selection, we randomly reduced the number of tactile scans from eight to seven. To establish ground truth for the computed transformation matrix, we manually aligned the tactile PC (full or partial) with the US template PC. As shown in Tab.~\ref{tab:sgg}, removing one tactile scan line does not significantly impact the final registration performance, indicating that slight variations in tactile scan paths have minimal effect on overall accuracy.

\begin{table}[ht]
\centering
\sisetup{
    table-number-alignment = center,
    table-figures-integer = 1,
    table-figures-decimal = 4
}
\caption{Registration Performance between Tactile and US PCs}
\label{table_registration_results}
\renewcommand\footnoterule{\kern -1ex}
\renewcommand{\arraystretch}{1.3}

\resizebox{0.4\textwidth}{!}{
    \begin{tabular}{lcc}
        \toprule
        \textbf{Positions} & \textbf{Transitional Error} & \textbf{Rotational Error} \\ 
        \midrule
        Position 1 & 3.38 mm & 0.28 def\\ 
        Position 2 & 3.58 mm& 0.58 deg \\ 
        \bottomrule
    \end{tabular}
}
\end{table}

\begin{table}[ht]
\centering
\caption{
Registration Performance under varying sizes of point cloud
}
\label{tab:sgg}
\begin{tabular}{ccccccc}
\toprule
\multirow{2}{*}{\textbf{Phantoms}} & \multicolumn{2}{c}{\textbf{Full PC}} & \multicolumn{2}{c}{\textbf{Partial PC 1}} & \multicolumn{2}{c}{\textbf{Partial PC 2}} \\
\cmidrule(lr){2-3} \cmidrule(lr){4-5} \cmidrule(lr){6-7} 
 & \textbf{Dist.} & \textbf{Ang.} & \textbf{Dist.} & \textbf{Ang.} & \textbf{Dist.} & \textbf{Ang.} \\
\midrule
Phantom B & 5.23 & 1.94 & 4.62 & 2.51 & 5.33 & 1.87 \\
Phantom C & 3.58 & 0.58 & 3.58 & 0.57 & 1.70 & 1.39\\
\bottomrule
\multicolumn{7}{l}{\textbf{Dist.}: distance error (mm), \textbf{Ang.}: angle error (deg).}
\end{tabular}
\end{table}










\begin{figure}[h]
\centering
\includegraphics[width=0.95\columnwidth]{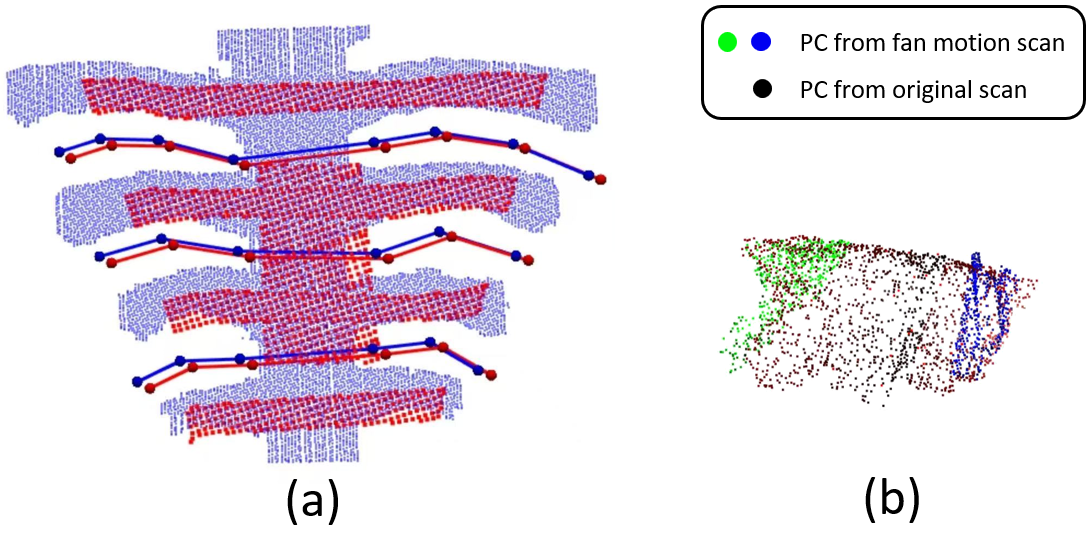}  
\caption{(a). The registration result of the US scanning path. The blue lines represent the ground truth scanning path, while the red lines represent the transformed scanning path. The blue point cloud is the transformed template US PC, and the red PC is the tactile PC. (b). US scan results using the transferred scanning path. Fan motion adjustments help to address incomplete regions of the scan target.}
\label{fig:reg_result}
\end{figure}


\subsection{Performance of Path Mapping \& Full Coverage Scanning}
\par
This section detailed the intercostal path mapping performance. An example is demonstrated in \revision{Fig.~\ref{fig:reg_result}}. To evaluate the accuracy of the scanning path transformation, we selected three different scanning paths along the rib gaps from the template US point cloud. We then applied both the ground truth transformation and our computed transformation to map these paths to the current position. In \revision{Fig.~\ref{fig:reg_result}}, the blue path represents the ground truth scanning path, while the red path corresponds to the transformed path obtained through our tactile registration. The Mean Nearest Neighbor Distance (MNND) error for all key waypoints is 3.41 mm, and the Hausdorff distance is 3.65 mm, demonstrating the accuracy of our method. Then, after performing the scan, we computed the volume of the \revision{pork meat} of interest from the US scan. Comparing this with the the volume ground truth obtained from CT scan, the MNND and HD are 0.69~mm and 2.2 mm, respectively.




\section{Conclusion}
\label{sec:conclusion}
In this study, we explored the potential of tactile cues as an alternative to ultrasound for bone surface point cloud extraction. By interpolating sparse tactile data, we generated a denser surface point cloud, which was then utilized in a registration algorithm to align preoperative scan paths with the current setting. Bone detection and registration were thoroughly validated across three distinct phantoms, demonstrating that even with sparse tactile data, our method achieved precise registration, with an MNND error of 3.41 mm and a Hausdorff distance of 3.65 mm. These findings confirm that tactile-based bone detection and scan path optimization provide a viable alternative to ultrasound imaging, with promising applications in robotic-assisted scanning and automated intercostal imaging. Future work will extend this approach to clinical settings and integrate it with B-mode ultrasound imaging to improve robustness and adaptability in characterizing subcutaneous bone surfaces.

\bibliographystyle{IEEEtran}
\balance
\bibliography{IEEEabrv,references}

\end{document}